\documentclass[11pt]{article}
\usepackage{amsmath,amsfonts,amsthm,amssymb}
\usepackage{url}

\newcommand{\data}{\mathcal{D}}
\newcommand{\gauss}{\mathcal{N}}
\usepackage{graphicx}
\usepackage{geometry}
\usepackage{epsfig}

\usepackage{amsthm}
\usepackage{amsmath}
\usepackage{amsfonts}
\usepackage{amssymb}
\usepackage{algorithm}
\usepackage{algorithmic}
\usepackage{color}
\usepackage{enumerate}
\usepackage{wrapfig}
\usepackage{subfigure}
\usepackage{bm}
\usepackage{tabu}
\usepackage{multirow}
\usepackage[english]{babel}
\usepackage{times}
\usepackage{graphicx}
\usepackage{url}
\usepackage[square,sort,comma,numbers]{natbib}

\newcommand{\argmax}{\operatornamewithlimits{argmax}}

\newtheorem{theorem*}{Theorem}

\usepackage{lineno,hyperref}
\begin{document}
\title{Fast Sampling Methods for
 Bayesian Max-margin Models}
\author{Wenbo Hu, Jun Zhu, Bo Zhang
\date{\today}
}
\maketitle

\begin{abstract}
Bayesian max-margin models have shown superiority in various practical applications, such as text categorization, collaborative prediction, social network link prediction and crowdsourcing, and they conjoin the flexibility of Bayesian modeling and predictive strengths of max-margin learning. However, Monte Carlo sampling for these models still remains challenging, especially for applications that involve large-scale datasets. In this paper, we present the stochastic subgradient Hamiltonian Monte Carlo (HMC) methods, which are easy to implement and computationally efficient. We show the approximate detailed balance property of subgradient HMC which reveals a natural and validated generalization of the ordinary HMC. Furthermore, we investigate the variants that use stochastic subsampling and thermostats for better scalability and mixing. Using stochastic subgradient Markov Chain Monte Carlo (MCMC), we efficiently solve the posterior inference task of various Bayesian max-margin models and extensive experimental results demonstrate the effectiveness of our approach.
\end{abstract}
\section{Introduction}
Bayesian max-margin~(BMM) models have been shown to be very effective in many real-world applications, such as text analysis~\citep{zhu2012medlda}, collaborative prediction~\citep{xu2012nonparametric}, social network link prediction~\citep{zhu2012max} and crowdsourcing~\citep{tian2015max}. Such BMM models conjoin the advantages of the discriminative max-margin learning and flexible Bayesian models, and they
achieve the best of the both worlds: obtaining the flexibility from a Bayesian model and meanwhile doing discriminative max-margin learning, through a newly-developed unified Bayesian inference framework, regularized Bayesian inference~(RegBayes)~\citep{zhu2014ilsvm}.

In order to deal with large-scale datasets, developing effective and scalable inference methods is a crucial problem for Bayesian max-margin models, which is becoming a norm in many application areas. Previous variational-approximation-based inference methods are raised to solve the BMM models with mean-field assumptions on posterior distributions~\citep{zhu2012medlda}. When the BMM models use nonparametric Bayesian priors, such variational methods need to adopt the model truncation to finish the variational approximation~\citep{zhu2011infinite,xu2013nonparametric}. Moreover, in such inference scheme, solving support vector machine (SVM) subproblems is time-consuming, which motivated the further developments of the Gibbs classifier formulation and the data augmentation-based Gibbs sampler~\citep{xu2013nonparametric,zhu14medlda,zhang2014max}.

In Bayesian inference, if we use a conjugate prior (w.r.t a given likelihood), %the posterior distribution is in the same family of the prior probability distribution, we call the prior and the likelihood are conjugate distributions. Given conjugate distributions,
we can easily derive the close-form posterior~\citep{gelman2014bayesian}. However, the BMM models are usually non-conjugate due to the non-smoothness of the hinge loss, which is often involved in an unnormalized pseudo-likelihood. The straightforward Gibbs sampler is not applicable due to the non-conjugacy. With a newly discovered data augmentation technique~\citep{polson2011data}, the augmented Gibbs sampler achieves accurate posterior sampling and is truncation-free for nonparametric BMM models~\citep{xu2013nonparametric,zhang2014max}. However, the Gibbs samplers with data augmentation are not efficient either in high-dimensional spaces as they often involve inverting large matrices~\citep{polson2011data}. Moreover, the benefit of introducing extra variables would be counteracted in the view of the extra computation on dealing with the extra sampling variables~\citep{roberts2002langevin}.

In this paper, we present the subgradient-based Hamiltonian Monte Carlo (HMC) methods for BMM models, which directly draw samples from the original posterior instead of the augmented one.
%After adopting some mild conditions of the posterior functions, we show that the subgradient-based HMC inherits the \emph{approximate detailed balance} property from the ordinary HMC.
After adopting some mild conditions of the posterior functions, we show the {approximate detailed balance} property for subgradient HMC methods.
Then using stochastic subgradient estimation~\citep{robbins1951stochastic,welling2011bayesian}, we further develop the stochastic subgradient MCMC for fast computation. By annealing the discretization stepsizes properly, our stochastic subgradient MCMC methods approximately converge to the target posteriors of basic Bayesian SVM fairly efficiently. To apply stochastic subgradient MCMC on two different types of BMM models with latent variables, we design two different inference algorithms for latent structure discovery, including a nonparametric Bayesian model. Our stochastic subgradient MCMC can achieve dramatically fast sampling and meanwhile draw accurate posterior samples. We carry out extensive empirical studies on large-scale applications to show the effectiveness and scalability of the presented stochastic subgradient MCMC methods for BMM models.

We note that there have been several previous attempts of using subgradient information in HMC or Langevin Monte Carlo~\citep{welling2011bayesian,neal2011mcmc}, yet our work stands as a first close investigation, in which we give the theoretical guarantee and carry out systematic studies on the stochastic subgradient MCMC for Bayesian max-margin learning.

\section{Preliminaries}\label{sec:background}
We first briefly review the Bayesian max-margin models with Gibbs classifiers. Then, we introduce the background knowledge of the inference methods, including Hamiltonian Monte Carlo (HMC) and its extension, as well as stochastic gradient Hamiltonian Monte Carlo.

\subsection{Bayesian Max-margin Models}
\label{sec:gibbs_classifier}
With the generic framework of \emph{RegBayes}~\citep{zhu2014ilsvm}, we can design more flexible Bayesian models by adding proper regularization on the target posterior.
Namely, after adding posterior regularization to a functional-optimization-reformulated Bayesian model, a \emph{RegBayes} model generally solves the following problem,
\begin{equation}\label{eqn:regbayes_bayes_max_margin}
\inf_{q(\mathcal{M})\in\mathcal{P}}~\mathrm{KL}\left(q(\mathcal{M})||\pi(\mathcal{M})\right)-\mathbb{E}_{q}[\log p(\data|\mathcal{M})]+c\cdot\mathcal{R}(q),
\end{equation}
where $\mathcal{M}$ denotes the model (parameters); $\mathcal{P}$ is the feasible space of probability distributions $q(\mathcal{M})$; $\mathrm{KL}\left(q(\cdot)||\pi(\cdot)\right)$ is the KL divergence from the target posterior $q(\mathcal{M})$ to the prior $\pi(\mathcal{M})$; $\data$ is the observation dataset; $c$ is a nonnegative regularization parameter and $\mathcal{R}(q)$ is a well-designed regularization term on $q$. It is not hard to show that if $c$ equals to $0$, the solution of problem~(\ref{eqn:regbayes_bayes_max_margin}) is the Bayes posterior $q(\mathcal{M}) \propto \pi(\mathcal{M}) p(\data|\mathcal{M})$. If $c$ is not zero, we have an extra dimension of freedom to introduce side information into the inference procedure through the posterior regularization term $\mathcal{R}(q)$. For example, when
the regularization $\mathcal{R}$ is defined as a hinge loss in supervised learning tasks, such \emph{Regbayes} models turn out to be Bayesian max-margin models and they successfully incorporate the flexibility of Bayesian models and the max-margin classifiers. This strategy has demonstrated promising performance in various tasks, including text classification and topic extraction~\citep{zhu2012medlda}, social network analysis~\citep{zhu2012max}, and matrix factorization~\citep{xu2012nonparametric}.

In this paper, we consider two examples of Bayesian max-margin models with latent variables, including \emph{max-margin topic model} (MedLDA)~\citep{zhu2012medlda} and \emph{infinite SVM} (iSVM)~\citep{zhu2011infinite}. But our methods can be applied to other BMM models. Specifically, MedLDA uses a topic model to find the latent topic representations of the documents and uses a max-margin classifier to do document classification. \emph{Infinite SVM} generally uses a Bayesian nonparametric Dirichlet process prior to describe data multi-modality and meanwhile uses max-margin classifiers to do discriminative tasks. More details of these two examples will be provided along the development of the proposed fast samplers for them.

\subsection{BMM models with a Gibbs classifier}\label{sec:BMM_Gibbs}
In the supervised learning setting, there are generally two types of classifiers that can be used with a Bayesian model to define a BMM model, namely, expected classifiers and Gibbs classifiers.
%For BMM models, we can construct two different formulations of inferring posterior distributions of classifiers: expected classifiers and Gibbs classifiers.
In this part, we give the introduction of the two formulations and analyze the merits of choosing Gibbs classifiers.

Let $\data=\{(x_{d},y_{d})\}_{d=1}^{D}$ be a given training set. For each data point $(x_d, y_d) \in \data$, $x_{d}$ denotes the input features and $y_{d}$ is the corresponding label, which can be binary or multi-valued. To build a classifier, a Bayesian max-margin model can either use the input features or learn a set of latent features. We use $x'_{d}$ to denote the features that are fit into a classifier. We consider the linear classifier parameterized by $\eta$. Then if the labels are binary, the prediction rule is defined as
\begin{equation}
\hat{y}_{d}={\rm sgn}\left[f(\eta, x'_{d})\right],~~f(\eta, x'_{d})=\eta^{\top}x'_{d},
\end{equation}
where ${\rm sgn}(\cdot)$ is the sign function.
%In different Bayesian max-margin models, the feature vector $x'_{d}$ could either be the learned latent feature vector or the raw feature vector. The weight vector $\eta_{d}$ is learned via max-margin learning.
%With $x'_{d}$ and $\eta_{d}$, we use the following discriminant function to do classification,
%For a single datum $x_{d}\in\data$ and its associated binary label $y_{d}$, a Bayesian max-margin model will learn its feature\junz{what's "correlated feature"? who says like this?} $x'_{d}$ through a flexible Bayesian model. The feature\junz{what's "released feature"? who says like this?} $x_d'$ could either be learned feature or raw feature, which is determined by the specific BMM model. For example, infinite SVM model uses the raw feature, while the MedLDA uses the learned latent topics as the input feature. Besides, the BMM model learns a weight vector $\eta_{d}$ via max-margin learning and uses the following discriminant function to do classification,
%\begin{equation}
%\hat{y}_{d}=f(\mathcal{M},x'_{d})={\rm sgn}[(\eta_{d})^{\top}x'_{d}].
%\end{equation}

For the above setting, an \emph{expected classifier} learns a posterior distribution $q(\eta)$ in a hypothesis space of classifiers that the $q$-weighted classifier $\hat{y}_{d}= \mathrm{sgn} \left(\mathbb{E}_{q}[f(\eta, x'_{d}]\right)$ will have the smallest possible risk, which is typically approximated by the training error $\mathcal{R}_{\data}(q)=\sum_{d=1}^{D}\mathbb{I}(\hat{y}_{d} \not= y_{d})$, where $\mathbb{I}(\cdot)$ is an indicator function that equals to 1 if predicate holds otherwise 0.
We define that $L(y_{d},\mathbb{E}_{q}[f(\eta, x'_{d})])=\max(0,l-y_{d}\mathbb{E}_{q}[f(\eta, x'_{d})])$ is the hinge loss function with regard to data point $d$ and $l(\geq1)$ is the cost of making a wrong prediction.
Then, we can use the \emph{RegBayes} formulation (Eqn.\ref{eqn:regbayes_bayes_max_margin}) to define a BMM model with an expected classifier by choosing the loss term $\mathcal{R}=\sum_{d=1}^{D}L(y_{d},\mathbb{E}_{q}[f(\eta,x'_{d})])$. It is known that the hinge loss $\mathcal{R}$ upper bounds the training error $\mathcal{R}_{\data}$.

Alternatively, the \emph{Gibbs classifier} draws a classifier $\eta$ according to $q(\eta)$ and uses it to do classification, which is proven to have nice generalization performance~\citep{mcallester2003pacbayes,germain2009pacbayes}.
In the Gibbs classifier, the corresponding loss is the \emph{expected hinge loss},
\begin{equation}
\mathcal{R}'=\sum_{d=1}^{D}\mathbb{E}_{q}[L(y_{d},f(\eta, x'_{d}))].
\end{equation}
Since the hinge loss function $L$ is convex, we can show that $\mathcal{R}'$ is an upper bound of $\mathcal{R}$, using Jensen's inequality:
\begin{equation}
\mathbb{E}_{q}[L(y_{d},f(\eta, x'_{d}))]\geq L(y_{d},\mathbb{E}_{q}[f(\eta, x'_{d})]).
\end{equation}
Then, the \emph{expected hinge loss} $\mathcal{R}'$ is also the upper bound of the expected training error of the Gibbs classifier
$\mathcal{R}'(q)\geq\sum_{d}\mathbb{E}_{q}[ \mathbb{I}(y_{d} \not= \hat{y}_{d})]$.
%So the loss term of a Gibbs classifier is a good surrogate loss. %Typical examples of BMM models with Gibbs classifier are also well studied, such as Gibbs MedLDA~\citep{zhu14medlda} and GiSVM~\citep{zhang2014max}.
Therefore, the \emph{Gibbs classifier} formulation gives a more relaxed model while at the same time can obtain uncertainty because we draw a single model for each time. In addition, with Gibbs classifiers, truncation-free sampling can be performed for BMM models with Bayesian nonparametric priors, which is more accurate than variational approximation. The BMM models with \emph{Gibbs classifiers} are already shown to have better performance of both classification results and efficiency of the inference algorithms~\citep{xu2013nonparametric,zhu14medlda,zhang2014max}.
%However, current MCMC inference methods for them need to use data augmentation technique, which involves large scale matrix inversion and extra computation on dealing with the auxiliary variables. So in this paper, we consider developing fast MCMC sampling methods for BMM models with Gibbs classifiers in large-scale settings.

\subsection{Hamiltonian Monte Carlo}
One popular MCMC inference method is Hamiltonian Monte Carlo (HMC), also known as Hybrid Monte Carlo~\citep{neal2011mcmc}.
Hamiltonian Monte Carlo is built on the molecular dynamics and the advantage of HMC over random walk Metropolis and Gibbs sampling is proposing	 a distant move with a high acceptance probability.
%\junz{say briefly about the advantages of HMC, compared to standard MCMC}.
%Hamiltonian Monte Carlo allows for global moves
%Hamiltonian Monte Carlo can propose to move to a distant point, with a high probability of acceptance by taking account of the .
More recently, the stochastic extensions of HMC are developed for fast sampling.
%\junz{silly error!!}	

Formally, we are interested in the posterior distribution $p(\theta | \data) \propto \exp( - U(\theta; \data) )$, where $\theta$ denotes the variables of interest and $U$ is the potential energy function in the Hamiltonian dynamics~\citep{arnol1989mathematical}.
Consider the general case where a posterior distribution jointly takes into account the prior belief and data. The energy function is written as
\setlength\arraycolsep{1pt} \begin{eqnarray}
U(\theta;\data) = -\log p_0(\theta) - \log p(\data | \theta), %\sum_{i=1}^N \log P(x_i | \theta),
\end{eqnarray}
where $p_0(\theta)$ is the prior and $p(\data | \theta)=\prod_{d}p(x_{d}|\theta)$ is the likelihood given the common \textrm{i.i.d} assumption\footnote{In the supervised learning setting, the likelihood should be $p(\data | \theta)=\prod_{d}p(x_{d},y_{d}|\theta)$.}.
%By convention, $\theta$ corresponds to the position variable $q$ in Hamiltonian dynamics.
After introducing auxiliary momentum variables $r$ and its symmetric positive-definite mass $M$, the HMC sampler simulates the joint distribution: $p(\theta,r) \propto \exp\left( - U(\theta; \data) -  r^\top M^{-1} r/2 \right)$.

Assuming a differentiable potential energy $U(\theta)$, we can use an HMC sampler to infer the posterior distribution via simulating the dynamics with some discretization integrators such as the Euler or leapfrog. Specifically, using the conventional leapfrog integrator with stepsize $h$, the HMC method performs the following steps:
 \begin{eqnarray}\label{eqn:HMC}
\left\{ \begin{array}{rl}
r_{t+\frac{1}{2}} &= r_{t}-\frac{h}{2} \nabla_\theta U(\theta_{t}|\data)   \\
\theta_{t+1} &= \theta_{t}+h M^{-1}r_{t+\frac{1}{2}}\\
r_{t+1} &= r_{t+\frac{1}{2}}-\frac{h}{2} \nabla_\theta U(\theta_{t+1}|\data),
\end{array}\right.
\end{eqnarray}
where $r_{0}$ is initialized as $r_{0}\sim\gauss(0,M)$.
Having obtained samples of $(\theta, r)$, we discard the momentum variable $r$ and get samples of $\theta$ from our target posterior.

In particular, if only one leapfrog step is used and $M$ is set to be the identity matrix, we can obtain Langevin Monte Carlo~(LMC), a special case of HMC~\citep{neal2011mcmc}.
%\junz{what do you mean by "unused"? "auxiliary"?}
%\wenbo{momentum is "auxiliary" in HMC. if do not use momentum(only one leapfrog step), HMC "degenerates" to langevin monte carlo}

To compensate for the discretization error, a Metropolis-Hastings correction step is employed to retain the invariance of the target distribution.

\subsection{Stochastic Gradient HMC}
One challenge of the gradient-based HMC methods on dealing with massive data is the expensive evaluation of the posterior gradient $\nabla_{\theta} U(\theta; \data)$. To save time, an unbiased noisy gradient estimate ${\nabla}_{\theta} \tilde{U}(\theta; \data)$ can be constructed by subsampling the whole dataset, as in stochastic optimization~\citep{robbins1951stochastic,bottou2010large}.

This idea was first proposed in~\citep{welling2011bayesian} to develop the stochastic gradient Langevin dynamics (SGLD), and was later extended by~\citep{chen2014stochastic} for stochastic gradient HMC with friction and by~\citep{ding2014bayesian} for stochastic gradient HMC with thermostats. In these stochastic MCMC methods, the gradient of the log-posterior is estimated as
\begin{equation}\label{eqn:noise gradient estimate}
{\nabla}_\theta\tilde{U}(\theta;\data) =  \frac{|\data|}{|\tilde{\data}|}\nabla_\theta  U(\theta;\tilde{\data}),
\end{equation}
where $\tilde{\data}$ is a randomly-drawn subset of $\data$. Since $|\tilde{\data}| \ll |\data|$, computing this noisy gradient estimate turns out much cheaper, hence rendering the overall algorithm scalable.

We now briefly review the stochastic gradient HMC with thermostats, or stochastic gradient Nos\'{e}-Hoover thermostat~(SGNHT)~\citep{ding2014bayesian}. SGNHT uses the simple Euler integrator and introduces a thermostat variable $\xi$ to control the momentum fluctuations as well as the injected noise.
The dynamics is simulated as:
\setlength\arraycolsep{1pt} \begin{eqnarray}\label{eqn:SGNHT}
\left\{ \begin{array}{rl}
r_{t+1} &= r_{t}-h\xi_{t} r_{t}-h{\nabla}_\theta \tilde{U}(\theta_{t}|\data)+\sqrt{2A}\gauss(0,h)\\
\theta_{t+1} &= \theta_{t}+h r_{t+1} \\
\xi_{t+1} &= \xi_{t}+h(\frac{1}{n}r^{\top}_{t+1}r_{t+1}-1),
\end{array}\right.
\end{eqnarray}
where $A$ is the diffusion factor parameter and $n$ is the dimension of $\theta$ and $r$.
$r_{0}$ is initialized from the standard normal distribution $\mathcal{N}(0,\mathbf{I})$ and $\xi_{0}$ is initialized as $A$.

Such stochastic gradient MCMC methods are shown to have a weak posterior-mean convergence instead of a strong sample-wise convergence~\citep{sato2014sgldconvergence,chen2015splitting}. Such weak convergence is sufficient in many real-world applications.

\section{Stochastic Subgradient MCMC}\label{sec:subgradient}
One central part in all the above HMC methods is the (stochastic) gradient of the log-posterior. However, such a gradient might not always be available. In this section, we investigate a more general subgradient-based HMC method, analyze its theoretical properties, and use it for the fast inference of Bayesian linear SVMs.

\subsection{Subgradient HMC and Its Approximate Detailed Balance}\label{sec:subgradienthmc}
When the log-posterior is non-differentiable, gradient-based HMC is not applicable. Using the more general subgradients could potentially address this problem, in analogy to the subgradient descent methods in deterministic optimization~\citep{shor1985minimization}.

By plugging the posterior subgradient $\partial_\theta U(\theta_{t}|\data)$ in the ordinary HMC, we come up with the subgradient HMC with a leapfrog method as:
\begin{eqnarray}\label{eqn:subgradient HMC}
\left\{ \begin{array}{rl}
r_{t+1/2} &= r_{t}-\frac{h}{2} \partial_\theta U(\theta_{t}|\data)   \\
\theta_{t+1} &= \theta_{t}+h M^{-1}r_{t+1/2}\\
r_{t+1} &= r_{t+1/2}-\frac{h}{2} \partial_\theta U(\theta_{t+1}|\data),
\end{array}\right.
\end{eqnarray}
where $r_{0}$ is initialized as $r_{0}\sim \gauss(0,M)$ and $h$ is the discretization stepsize.

%\subsection{Approximate Detailed Balance}\label{sec:approximate-db}
From a theoretical perspective, we may not be able to readily analyze the volume preservation property of the Hamiltonian dynamics with a non-differentiable potential energy nor the detailed balance of a general subgradient HMC sampler. Instead, we give an approximated theoretical analysis based on several practical assumptions of the potential energy.

In practical Bayesian models, the non-smoothness of the posterior often lies in the hinge loss induced likelihoods which are mainly considered in this paper. These posteriors are continuous everywhere and piece-wise smooth with only a finite number of non-smooth points. The sampler will hit those non-differentiable states with probability zero. Under such practical assumptions, we show the following \textit{approximate detailed balance} property, which claims that the subgradient HMC satisfies the detailed balance property with a polynomial smooth of the potential energy .
%subgradient HMC converges to an approximate posterior with detailed balance and the approximate posterior can be as close as possible to the true posterior.

We first give a polynomial smooth of the potential energy $U_{0}$.
The continuous and piece-wise differentiable posterior $U_{0}$ is non-smooth on a finite set $S=\{s_{i}\}_{i=1}^{m}$ and then the $\epsilon$-neighborhoods around all $s_{i}$ are defined as $B(s_{i},\epsilon)=\{\theta|\lVert\theta-s_{i}\rVert<\epsilon\},~i=1,2,\cdots,m.$
By setting $\epsilon$ small enough, the $\epsilon$-neighborhoods can be mutually disjoint: $B(s_{i},\epsilon)\cap B(s_{j},\epsilon)=\varnothing,~\forall s_i,s_j\in S,i \not= j.$ Using such mutually disjoint neighborhoods, $U_{\epsilon}$ will be constructed as
\begin{equation}\label{eqn:polynomial smooth}
U_{\epsilon}(\theta)= \left\{ \begin{array}{rl}
   U_0(\theta), & ~~~\forall s_i\in S,~\theta\not\in B(s_{i},\epsilon)\\
   \mathcal{P}_{i,\epsilon}(\theta), & ~~~\theta\in B(s_{i},\epsilon),	\end{array} \right.
\end{equation}
where $\mathcal{P}_{i,\epsilon}$ is a multi-dimensional Hermite's interpolating polynomial~\citep{bajaj1993multi} satisfying
\begin{eqnarray}\label{eqn:polynomial construction}
\left\{ \begin{array}{ll}
\mathcal{P}_{i,\epsilon}(s_{i}\pm\epsilon) &= U_{0}(s_{i}\pm\epsilon), \\
\nabla_{q}\mathcal{P}_{i,\epsilon}(s_{i}\pm\epsilon) &= \partial_{q}U_{0}(s_{i}\pm\epsilon).
\end{array} \right.
\end{eqnarray}
%The polynomial smooth is exact except the $\epsilon$-neighbor of the non-smooth points and moreover, such $\epsilon$-neighbor can be arbitrarily small.
According to the definition of $U_\epsilon$~(Eqn.~\ref{eqn:polynomial smooth},~\ref{eqn:polynomial construction}), we can see that $U_{\epsilon}$ is smooth everywhere. Moreover, when $\theta\not\in B(s_{i},\epsilon), \forall s_i\in S$, we have
\begin{equation}\label{eqn:converginggradient}
U_{\epsilon}(\theta) = U_{0}(\theta)~,~\nabla_{\theta}U_{\epsilon}(\theta) = \partial_{\theta}U_{0}(\theta).
\end{equation}
%when $\theta\not\in B(s_{i},\epsilon), \forall s_i\in S$.

%When drawing finite number of samples from $U_{0}$, the sampler will hit $S$ with probability zero. %, by setting $\epsilon$ goes to zero. %When $\epsilon \rightarrow 0$, we have $\cup_{i}B(s_i,\epsilon)\rightarrow S$.
%Using the polynomial smooth, the posterior subgradients $\partial_{\theta}U_{0}$ used in Eqn.~\ref{eqn:subgradient HMC} is approximately the same with $\nabla_{\theta}U_{\epsilon}$.
%Considering that $U_{0}$ is continuous everywhere and the gradient of $U_{0}$ is continuous except the finite set $S$, we have
%\begin{equation}\label{eqn:polynomial convergence}
%\lim_{\epsilon\rightarrow 0}U_{\epsilon}(\theta) = U_{0}(\theta)~,~\lim_{\epsilon\rightarrow 0}\nabla_{\theta}U_{\epsilon}(\theta) \overset{\text{a.e.}}{=} \partial_{\theta}U_{0}(\theta).
%\end{equation}
%The polynomial smooth Eqn.~\ref{eqn:polynomial smooth} is not unique but we can always let $\epsilon$ be very small.
When $\epsilon$ is small enough, the posterior subgradients $\partial_{\theta}U_{0}$ used in Eqn.~\ref{eqn:subgradient HMC} is approximately the same with $\nabla_{\theta}U_{\epsilon}$ and it will be scarcely possible for the sampler to hit those neighborhoods since the measure of the neighborhood is bounded by $\epsilon$.
 Then subgradient HMC can be equivalent to drawing samples from a smooth posterior $U{\epsilon}$ instead. With this approximation, the subgradient HMC satisfies detailed balance and is thus valid for generating approximate samples from the true posterior $U_{0}$.

\begin{wrapfigure}{r}{0.4\textwidth}
\vspace{-20pt}
  \begin{center}
    \includegraphics[width=0.4\textwidth]{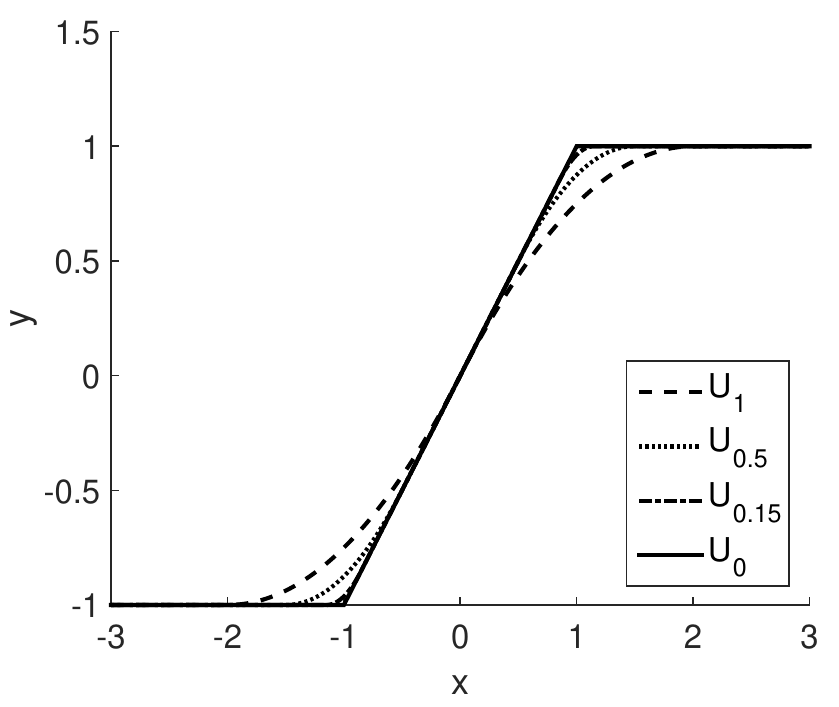}%{./Pictures/mainscreen1.png}
     \vspace{-10pt}
    \caption{Illustration of Polynomial Smooth Construction}
    \label{fig:approximation demo}
  \end{center}
  \vspace{-20pt}
  \vspace{0pt}
\end{wrapfigure}

We give an intuitive illustration of the theoretical analysis. In Fig.~\ref{fig:approximation demo}, we construct several polynomial smooth functions $U_{\epsilon}$ for a continuous but non-smooth function $U_{0}$. As can be seen, when $\epsilon$ is as small as 0.15, $U_{0.15}$ is very close to $U_{0}$ and it's very unlikely for a sampler to use finite samples (such as 100 samples), to hit the two neighborhoods $B(-1,0.15)$ and $B(1,0.15)$.

%Finally, we note that the construction is not unique and the smooth is more accurate if the $\epsilon$ is smaller. We can always let $\epsilon$ go to zero and then the approximate posterior $.

\subsection{Stochastic Subgradient MCMC in Practice}
We can obtain the version of stochastic subgradient Langevin dynamics~(SSGLD) by replacing the gradient of the log-posterior with its subgradient. More formally, SSGLD generates samples by simulating the following dynamics:
\begin{eqnarray}\label{eqn:SSGLD}
\left\{ \begin{array}{rl}
\theta_{t+1} &= \theta_{t}-\frac{h^{2}}{2}\partial_\theta \tilde{U}(\theta_{t+1}|\data)+h \nu_t \\
\nu_{t} &\sim  \gauss(0,I),
\end{array}\right.
\end{eqnarray}
where $\partial_\theta \tilde{U}(\theta;\data) \triangleq  -\partial_\theta \log p(\theta)-\frac{|\data|}{|\tilde{\data}|}\partial_\theta \log p(\tilde{\data}|\theta)$ is the stochastic noisy estimate of the subgradient $\partial_\theta U(\theta;\data)$.

In existing SGLD methods~\citep{welling2011bayesian}, it is recommended to use a polynomial decaying stepsize to save the MH correction step of the Langevin proposals. When the stepsize properly decays, the Markov chain would gradually converge to the target posterior.
One subtle part of the method is thus on tuning the discretization stepsize. A pre-specified annealing scheme (if not chosen properly) would make the chain either miss or oscillate around the target. More recent work~\citep{TehThiVol2014a} recommends some relatively optimal scheme for SGLD. Inspired by adaptive stepsizes for (sub)gradient descent (AgaGrad) methods~\citep{duchi2011adaptive}, we, in this paper, adopt the same adaptive stepsize setting for our SSGLD methods~\citep{li2015preconditioned}. As we shall see in the experiments, such a scheme is beneficial to yield faster mixing speeds.

We can derive stochastic subgradient Hamiltonian Monte Carlo likewise.
We adopt an improved version of stochastic gradient HMC~\citep{ding2014bayesian} to derive our stochastic subgradient Nose-Hoover thermostat (SSGNHT),
which generates samples via the following iterations:
 \begin{eqnarray}\label{eqn:SSGNHT}
\left\{ \begin{array}{rl}
r_{t+1} &= r_{t} - \xi_t r_{t}h - h\partial_\theta \tilde{U}(\theta_{t}|\data)+\sqrt{2A} \gauss(0,h)\\
\theta_{t+1} &= \theta_{t}+h r_{t+1} \\
\xi_{t+1} &= \xi_{t}+h(\frac{1}{n}r^{\top}_{t}r_{t}-1).
\end{array}\right.
\end{eqnarray}
%Here, $n$ is the dimension of $\theta$.
Again we omit the MH correction step and the SSGNHT simulations would generate posterior samples more efficiently with the properly decaying stepsizes and thermostat initialization.
\subsection{Stochastic Subgradient MCMC for Bayesian Linear SVMs}
The stochastic subgradient MCMC can be used for fast sampling of Bayesian linear SVM. Let $\data = \{(x_d, y_d)\}_{d=1}^{D}$ be the given training dataset, where $x_d$ is the $n$-dimensional feature vector of the $d$-th instance and $y_d \in \{-1,+1\}$ is the binary label.
We use linear classifiers with a weight vector $\eta\in\mathbb{R}^n$ and the decision rule is naturally $\hat{y} = \textrm{sgn}(\eta^\top x)$. Then for a Bayesian linear SVM model, we are interested in learning the posterior distribution $p(\eta | \data) \propto p_0(\eta)\prod_{d} \psi(y_d | x_d, \eta)$. The prior is commonly set as a standard normal distribution $p_{0}(\eta) = \gauss(0, I)$, and the per-datum unnormalized likelihood is $\psi(y_d | x_d, \eta) = \exp(-c \cdot \max(0,l-y_{d}\eta^{\top}x_{d}))$.
%Then the log-posterior subgradient for Bayesian linear SVM can be estimated as
%\junz{you use both sgn and sign!}
Then, the subgradient of the log-posterior involves evaluating the subgradient of the non-differentiable log-likelihood
\begin{eqnarray}\label{eqn:subgradient for classifier}
\partial_\eta \log \psi (y_d | x_d, \eta) =
\begin{cases}
- c y_{d}x_{d} & l-y_{d}\eta^{\top}x_{d} > 0  \\
0 & l-y_{d}\eta^{\top}x_{d} \leq 0.
\end{cases}
\end{eqnarray}
With this subgradient, we can use the stochastic subgradient MCMC method to do fast sampling for the Bayesian linear SVM model.

\section{Fast Sampling for Bayesian Max-margin Models with Latent Variables}\label{sec:sampling_mixture_of_svms}
%\junz{Bayesian linear SVM is also a BMM model. This title and the following statements should be changed to reflect the difference. Call them BMM models with latent variables?}

We now show how to leverage the above stochastic subgradient MCMC methods to derive fast sampling algorithms for Bayesian max-margin models with latent variables. We develop algorithms
%\junz{paradigm is a very strong word. do you mean "algorithm"?}
for two different BMM models with latent variables.

\subsection{Fast Sampling for Max-margin Topic Models}
%Another type of Bayesian max-margin learning model does not consider Bayesian non-parametric priors, so the number of the model number is fixed. Then we needn't use a Gibbs sampler to sample the model number and instead we can use stochastic subgradient MCMC to directly sample the posterior. Compared with the HMC-within-Gibbs structure before, this is fully stochastic subgradient sampler which is much faster the previous HMC-within-Gibbs structure. In this part, we use Gibbs MedLDA~\citep{zhu14medlda}(see Fig.~\ref{fig:medlda graphical}) as an example.

For parametric BMM models, whose model parameter number is fixed,
we just calculate the (stochastic) log-posterior subgradient and run our stochastic subgradient MCMC method. In this part, we use Gibbs MedLDA~\citep{zhu14medlda} as an example to show how to do fast sampling for parametric BMM models.

\begin{figure}[ht]\vspace{-.1cm} \centering
{
\includegraphics[width=.7\columnwidth]{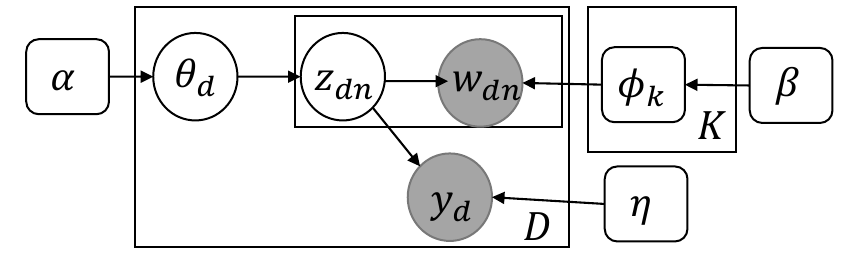}
}
\vspace{-10pt}
\caption{Graphical model representation of Gibbs MedLDA}
\vspace{-10pt}
\label{fig:medlda graphical}
\end{figure}
\subsubsection{Gibbs MedLDA}

As illustrated in Fig.~\ref{fig:medlda graphical}, the max-margin topic model has two parts: 1) a latent Dirichlet allocation model for modeling underlying topic structures of the given documents and 2) a max-margin classifier for predicting document labels. The LDA part is a hierarchical Bayesian model which uses an admixture of $K$ topics, $\mathbf{\Phi}=\{\mathbf{\Phi}_{k}\}_{k=1}^{K}$, as a latent document representation. Here each topic $\mathbf{\Phi}_k$ is a multinomial distribution over a $V$-word vocabulary and has the symmetric Dirichlet prior $\mathrm{Dir}(\beta)$.
%Each document has a topic proportion $\theta$ with the symmetric Dirichlet prior $\mathrm{Dir}(\alpha)$.
%The generative process of a single document $d$ is
For a single document $d$, $N_d$ words are generated and the detailed process is
\begin{enumerate}\setlength{\itemsep}{-0.1cm}
\item draw a topic proportion $\theta_{d}\sim \mathrm{Dir}(\alpha),$
\item for each word $n~(1\leq n \leq N_d)$:
      \begin{description}\setlength{\itemsep}{-0.1cm}
            \item[(a)]  draw a topic assignment $z_{dn}\sim \mathrm{Multinomial}(\theta_{d}),$
            \item[(b)] draw the observed word $w_{dn}\sim \mathrm{Multinomial}(\Phi_{z_{dn}}).$
      \end{description}
\end{enumerate}
Given a set of documents $\mathbf{W}=\{w_{d}\}_{d=1}^{D}$, we denote its latent topic proportions as $\mathbf{\Theta}=\{\theta_{d}\}_{d=1}^{D}$ and  its topic assignments as $\mathbf{Z}=\{z_{d}\}_{d=1}^{D},z_d=\{z_{dn}\}_{n=1}^{N_d}$. Let $\bar{z}_{d}$ be the average topic assignments of the words in document $d$, with element $\bar{z}_{dk}=\frac{1}{N_{d}}\sum_{n}^{N_d}\mathbb{I}(z_{dn}=k)$.

We use the Gibbs classifier formulation to build the Gibbs MedLDA model. If we have drawn a sample of the topic assignments $\mathbf{Z}$ and the classifier weights $\eta$ from the posterior distribution, we can get the prediction of the document label $y_{d}\in\{1,2,\cdots,L\}$ as,
\begin{equation}
\hat{y}_{d}=\argmax_{y}f(y,\bar{z}_{d}|\eta)~~~~f(y,\bar{z}_{d}|\eta)=\eta^{\top}g(y,\bar{z}_{d}),~y\in\{1,2,\cdots,L\},
\end{equation}
where $g(y,\bar{z}_d)$ is a long vector consisting of $L$ subvectors with the $y$-th being $\bar{z}_d$ and all others being zero.
%Also we use RegBayes to incorporate the supervised information and the Gibbs classifier with expected hinge loss to get uncertainty.
The corresponding \emph{expected hinge loss} is
\begin{equation}\label{eqn:multiclass for medlda}
\mathcal{R}'\left(q(\eta,\mathbf{\Theta},\mathbf{Z},\mathbf{\Phi})\right)=\sum_{d=1}^{D}\mathbb{E}_{q}\left[\max\left(0,l+\max_{y\neq y_d}f(y,\bar{z}_{d}|\eta)-f(y_{d},\bar{z}_{d}|\eta)\right)\right].
\end{equation}
Then, Gibbs MedLDA infers the latent topic assignments $\mathbf{Z}$ and the classifier weights $\eta$ by solving the following \emph{RegBayes} problem:
\begin{equation}
\min_{q(\eta,\mathbf{\Theta},\mathbf{Z},\mathbf{\Phi})}\mathcal{L}\left(q(\eta,\mathbf{\Theta},\mathbf{Z},\mathbf{\Phi})\right)+c\cdot\mathcal{R}'\left(q(\eta,\mathbf{\Theta},\mathbf{Z},\mathbf{\Phi})\right),
\end{equation}
where $\mathcal{L}=\mathrm{KL}\big(q||p_{0}(\eta,\mathbf{\Theta},\mathbf{Z},\mathbf{\Phi})\big)-\mathbb{E}_{q}\big[\log(p(\mathbf{W}|\mathbf{Z},\mathbf{\Phi})\big]$ is the reformulated objective when doing standard Bayesian inference.

\subsubsection{Fast Sampling for Gibbs MedLDA}
%The semi-collapsed distribution is considered (only $\mathbf{\Theta}$ collapsed, compared with collapsed Gibbs sampler~\citep{griffiths2004finding} which collapsed both $\mathbf{\Phi}$ and $\mathbf{\Theta}$),
Instead of sampling in the whole space, which may lead to low efficiency~\citep{griffiths2004finding}, we collapse out $\mathbf{\Theta}$ and draw samples form the collapsed distribution,
\begin{equation}
\nonumber p(\mathbf{W},\mathbf{Z},\mathbf{\Phi},y|\alpha,\beta)   =   p(\eta)p(\mathbf{\Phi}|\mathbf{\beta})\prod_{d=1}^{D}p(\mathbf{w}_{d},z_{d}|\alpha,\mathbf{\Phi})\psi(y_{d}|z_{d},\mathbf{\eta}),
\end{equation}
where
\begin{equation}
p(\mathbf{w}_d,z_{d}|\alpha,\mathbf{\Phi})=\prod_{k=1}^{K}\frac{\Gamma(\alpha+C_{dk\cdot})}{\Gamma(\alpha)}\prod_{w=1}^{W}\Phi_{kw}^{C_{dkw}}.
\end{equation}
$C_{dk\cdot}$ is the number of words in document $d$ that is assigned to topic $k$ and $C_{dkw}$ is the number of words $w$ in document $d$ that is assigned to topic $k$. $\psi(y_{d}|z_{d},\mathbf{\eta})$ is defined as,%The hinge loss with regard to the document $d$ is
\begin{equation}
\psi(y_{d}|z_{d},\mathbf{\eta})=\exp\left[-c\max\left(0,l+\max_{y\neq y_d}\eta^{\top} g(y,\bar{z}_{d})-\eta^{\top} g(y_{d},\bar{z}_{d})\right)\right].
\end{equation}

For the collapsed posterior of MedLDA, we can sample classifiers $\eta$ using stochastic subgradient MCMC and sample the topic model parameters $\mathbf{\Phi}$ using the SGRLD method~\citep{patterson2013stochastic}.
With the randomly-drawn document minibatch $\tilde{\mathbf{W}}$, we get the stochastic subgradient of the log posterior with respect to $\eta$ as,
 \begin{eqnarray}\label{eqn:stochastic_subgradient_eta}
\left\{ \begin{array}{rl}
\partial_{\eta}\log\psi&=0; ~~\text{if} ~\psi(y_{d}|z_{d},\eta)=1, \\
\partial_{\eta_{y^{*}}}\log\psi&=-c\bar{z}_{d},\partial_{\eta_{y_{d}}}\log\psi=c\bar{z}_{d};~ \text{if}~\psi(y_{d}|z_{d},\eta)<1,
\end{array}\right.
\end{eqnarray}
%equals $0$, if $l+\max_{y\neq y_d}\eta^{\top} g(y,z_{d})\leq\eta^{\top} g(y_{d},z_{d})$ and otherwise, the subgradients of $\psi$ over $\eta$ is
%\begin{eqnarray}\label{eqn:subgradient multiclass hinge loss}
%\partial_{\eta_{\hat{y}}}\psi(y_{d},x_{d},\eta)=-c\bar{z}_{d},~~~
%\partial_{\eta_{y_{d}}}\psi(y_{d},x_{d},\eta)=c\bar{z}_{d}
%\end{eqnarray}
where $y^{*}=\argmax_{y\not=y}\eta^{\top}g(y,\bar{z}_{d})$. Here, $\eta_{y}$ is the $y$-th subvector of $\eta$ which is corresponding to the non-zero elements of $g(y,\bar{z}_{d})$ and in the second case of the calculation, the subgradients with respect to the unmentioned subvectors of $\eta$ are zero. With the stochastic posterior subgradient with respect to $\eta$, we can use stochastic subgradient MCMC to sample $\eta$.

 We use the expanded-mean formulation for $\Phi$:
$\Phi_{kn}={|\pi_{kn}|}/{\left(\sum_{n}|\pi_{kn}|\right)}$ and follow the SGRLD iterations to sample the admixture $\Phi$ on the Riemannian manifold~(Eqn.~10 in~\citep{patterson2013stochastic}).
  % Here only the $y_{d}$-th and $\hat{y}_{d}$-th subvector of $\eta$ has nonzero related subgradients and subgradients over other subvectors of $\eta$ are zero. Specially, if the prediction of $y_d$ is correct $\hat{y_d}=y_d$ but

The stochastic posterior (sub)gradients with respect to $\Phi$ and $\eta$ are calculated given the expectation of $\bar{z}$~\citep{mimno2012sparseonlinelda}.
To calculate the expectation of $\bar{z}$, the Gibbs sampling iterations for the topic assignments of document $d$ is as follows:
\begin{equation}\label{eqn:gibbs_singledocument}
p(z_{dn}=k|z_{d,-n},\Phi,\eta)\propto(\alpha+C_{dk\cdot}^{-n})\Phi_{kn}\psi(y_{d}|\bar{z}^{*}_{d},\mathbf{\eta}),
\end{equation}
where $z_{d,-n}$ is the topic assignments of other documents, $\bar{z}^{*}_{d}$ is the average topic assignments $\bar{z}_d$ after setting topic $z_{dn}$ as $k$ and $C_{dk\cdot}^{-n}$ is the number of words assignment as topic $k$ in document $d$ after removing word $n$.
With the learned topic admixture $\mathbf{\Phi}$ and classifier weights $\eta$, we randomly draw a sample of $\mathbf{\Phi}$ and $\eta$ and make predictions as described in~\citep{zhu14medlda}.
The overall stochastic sampler for Gibbs MedLDA is concluded in Algorithm~\ref{alg:sgrld_medlda}.
%At  each iteration, we use Gibbs sampler to get the expectation of $\bar{z}$, and use a single sample of $\bar{z}$ to update $\eta$ and $\Phi$.
 \begin{algorithm}[H]
   \caption{SSGRLD For Gibbs MedLDA}
   \label{alg:sgrld_medlda}
\begin{algorithmic}
   \STATE {\bfseries Input:} documents $(w_{d},y_{d}),d=1,\cdots,D$.
   \STATE Initialization
   \REPEAT
   \STATE Draw a stochastic subset $\tilde{\data}$
   \STATE Draw topic assignments of the documents in $\tilde{\data}$ using Eqn.~\ref{eqn:gibbs_singledocument}
   \STATE Compute stochastic posterior (sub)gradient with respect to $\Phi$ and $\eta$
   \STATE Run subgradient sampler for $\eta$ and $\Phi$ with the stochastic posterior subgradient
   \UNTIL{Converge}
\end{algorithmic}
\end{algorithm}

\subsection{Fast Sampling for Infinite SVMs}
Another important type of Bayesian max-margin models with latent variables uses Bayesian nonparametric priors. Such BMM models are defined on infinite-dimensional spaces and the size of the models will be learned from the data. Typical example of this type is infinite SVM~\citep{zhu2011infinite} and we use the HMC-within-Gibbs strategy to build fast sampling methods for this type of models.
%Here, we show how to do fast sampling for GiSVM using HMC-within-Gibbs strategy.
%We use the HMC-within-Gibbs strategy for GiSVM: sample the component assignments using Gibbs sampler and then given the component assignments, use fast sampling methods to infer other model parameters.

%It is difficult to use stochastic subsampling technique to infer the unbiased samples of the model numbers, such as the cluster number in Dirichlet process mixture model.

\subsubsection{Gibbs infinite SVM}

\begin{wrapfigure}{r}{0.4\textwidth}  \vspace{-60pt}
  \begin{center}
    \includegraphics[width=0.4\textwidth]{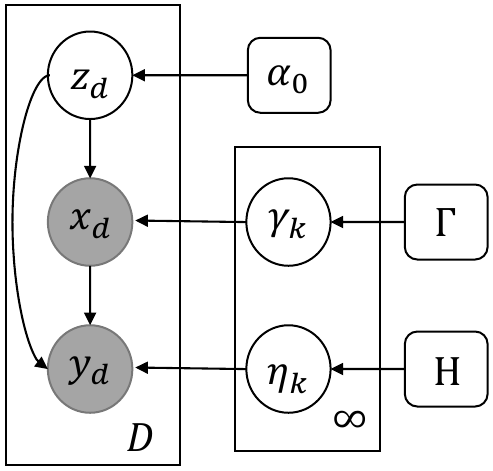}  \vspace{-20pt}
    \caption{Graphical model representation of Gibbs iSVM}
    \label{fig:isvm graphical}
  \end{center}
  \vspace{-30pt}
\end{wrapfigure}
Real world data often have some latent clustering structures, where mixture-of-experts models are generally capable of capturing these local structures. When each expert is a linear SVM, the resultant mixture of SVMs learns a non-linear model instead of simply a linear one~\citep{collobert2002parallel,fu2010mixing}. Recent work further presents a nonparametric extension, infinite SVM (iSVM)~\citep{zhu2011infinite} (See Fig.~\ref{fig:isvm graphical}), which automatically infers the number of experts. Below, we apply the subgradient-based fast sampling method to infinite SVM.

Given a set of data $\data=\{(x_{d},y_{d})\}_{d=1}^{D}$, we let $z_d$ denote the component assignment for the datum $x_d$. Each component is associated with a linear classifier $\eta_{z_{d}}$ and a Gaussian likelihood $(\mu_{z_{d}}, \Sigma_{z_{d}})$ to describe the input features.\footnote{The Gaussian likelihood is optional.}
All the parameters follow some priors: a standard Gaussian prior for $\eta$ and a Gaussian-Inverse-Wishart conjugate prior for $(\mu, \Sigma)$. In iSVM, we choose a \emph{Chinese Restaurant Process}~(CRP)~\citep{pitman2006combinatorial} prior for $Z$.
%With such Bayesian nonparametric prior, we learn the number of the clusters from the data and do not need model selection.

Though alternative approaches exist, we define the expert classifier as a Gibbs classifier to get uncertainty for the assignments $Z$ and the classifier weights $\eta$. Namely, given the posterior distribution $q(Z, \eta)$, the Gibbs classifier draws a component assignment $z_d$ and a classifier $\eta_{z_d}$ for each data point $x_d$ and makes prediction:
\begin{equation}
\hat{y}_{d}=\argmax_{y}f(y,x_{d},z_{d})~~~~f(y,x_{d},z_{d})=\eta_{z_d}^{\top}g(y,x_{d}),~y\in\{1,2,\cdots,L\},
\end{equation}
where $g(y,x_{d})$ is a long vector consisting of $L$ subvectors with the $y$-th being $x_d$ and all others being zero.
We adopt the expected hinge loss for Gibbs iSVM,
% (Eqn.~\ref{eqn:multiclass for medlda}),
\begin{equation}
\mathcal{R}'(q(\eta,\gamma,Z))=\mathbb{E}_{q(Z,\eta,\gamma)}\left[\sum_{d=1}^{D}\max\left(0,l+\max_{y\neq y_d}f(y,x_{d},z_{d})-f(y_d,x_{d},z_{d})\right)\right].
\end{equation}
Together with the Gibbs classifier and the expected hinge loss, we can define a \emph{RegBayes} model for the mixture of Gibbs classifiers:
\begin{equation}
\min_{q(Z,\eta,\gamma)}\mathcal{L}(q(Z,\eta,\gamma))+c\cdot\mathcal{R}'(q(Z,\eta,\gamma)),
\end{equation}
where $\gamma=(\mu,\Sigma)$ are the mean and variance parameters for each Gaussian component and $\mathcal{L}=\mathrm{KL}(q||p_{0}(\eta,\gamma,Z)-\mathbb{E}_{q}\big[\log(p(X|Z,\gamma)\big]$ is the objective function when doing standard Bayesian inference.
%ss
%Existing Gibbs sampling methods adopt the data augmentation technique~\citep{polson2011data} to perform a Gibbs sampler for the following posterior distribution
With regard to the \emph{RegBayes} formulation in Eqn.~\ref{eqn:regbayes_bayes_max_margin}, the normalized posterior distribution of infinite SVM is
\begin{equation}\label{eqn:posterior_isvm}
q(Z,\eta,\gamma)\propto p_{0}(\eta,\gamma,Z)p(X|Z,\gamma)\prod_{d=1}^{D} \psi(y_d | x_d, \eta_{z_d}),
\end{equation}
where $\psi(y_d | x_d, \eta_{z_d})=\exp(-c\max\left(0,l+\max_{y\neq y_d}f(y,x_{d},z_{d})-f(y_{d},x_{d},z_{d})\right))$.
We refer readers to~\citep{zhu2011infinite,zhang2014max} for more details.

\subsubsection{Fast sampling for Gibbs iSVM}
We develop the fast sampling method for Gibbs iSVM by incorporating the stochastic subgradient MCMC method within the loop of a Gibbs sampler. The HMC-within-Gibbs strategy for iSVM is detailed below.
%naturally replacing the Gibbs sampling $\eta$ step with stochastic subgradient MCMC.
%In the Gibbs sampler, with Gaussian component parameters $\gamma$ collapsed, the component assignments $Z$ of each datum is sampled given $\eta$ and then given $Z$ the classifiers $\eta$ is sampled using stochastic subgradient MCMC.
%Now we give the details.
% of fast sampling for Gibbs iSVM.

\textbf{For Z}: Give $\eta$, the conditional distribution is
\begin{equation}
p(Z|\eta)\propto p_{0}(Z)p(X|Z)\psi(Y|Z,\eta),
\end{equation}
where $p(X|Z)=\int p_{0}(\gamma)p(X|Z,\gamma)\mathrm{d}\gamma$ is the marginal distribution via collapsing $\gamma$ and $p_{0}(Z)$ is the CRP prior.
%Then, sampling $Z$ step which has two cases for a single data point's cluster assignment.
Let $\alpha_{0}$ be the hyper-parameter of the CRP prior and $n_{-d,k}$ be the number of data points that belong to component $k$ except $d$. Given classifiers $\eta$ and assignments of other data points $Z_{-d}$, we sample component assignments $z_{d}$ by normalizing the following two probabilities (existing component $k$ and a new component):
\begin{description}
  \item[1)] $p(z_{d}=k|Z_{-d},\eta)\propto n_{-d,k}\psi(y_{d}|z_{d}=k,\eta_{k})\cdot p(x_{d}|Z_{-d},X_{-d}^{k})$
  \item[2)] $p(z_{d}=\emph{new}|Z_{-d},\eta)\propto \alpha_{0}p(x_{d})\int\psi(y_{d}|\eta')p_{0}(\eta')\mathrm{d}\eta'$
\end{description}
%The above probability can be computed in the closed form or via importance sampling approximation~\citep{zhang2014max}.
In case 2), $p(x_{d})=\int p(x_{d}|\gamma)p_{0}(\gamma)\mathrm{d}\gamma$ is the likelihood of the data $d$ and can be computed in closed-form using the conjugate property. The second integral in case 2) can be approximated by using importance sampling.

\textbf{For $\eta$}: Give $Z$, the number of active cluster is known. We need to efficiently sample the classifier weights $\eta_{k}$ of each component $k$ from the following conditional distribution,
\begin{equation}\label{eqn:posterior_isvm_eta}
p(\eta_{k}|Z)\propto p_{0}(\eta_{k})\prod_{d:z_{d}=k}\psi(y_{d}|z_{d},\eta_{z_d}),
\end{equation}
where $p_{0}(\eta_{k})$ is a standard normal prior. With our proposed stochastic subgradient MCMC, the classifiers $\eta$ can be directly sampled using only a minibatch of whole dataset. Here, we give the stochastic subgradients of the log conditional distribution:
\begin{equation}
{\partial}_{\eta_{k}}\log \left[p_{0}(\eta_{k})\prod_{d:z_{d}=k}\psi(y_{d}|z_{d},\eta_{z_d})\right]\approx-\eta_{k}+
\frac{|\data|}{|\tilde{\data}|}\sum_{d:z_{d}=k,(x_{d},y_{d})\in\tilde{\data}}\partial_\eta \log \psi (y_d | x_d, \eta_{z_d}),
\end{equation}
where the subgradients of the multi-class hinge loss $\psi (y_d | x_d, \eta_{z_d})$ are similarly defined as Eqn.~\ref{eqn:stochastic_subgradient_eta}.
%: $\partial_{\eta}\psi(y_{d},x_{d},\eta)$ equals $0$, if $l+\max_{y\neq y_d}\eta^{\top} g(y,z_{d})\leq\eta^{\top} g(y_{d},z_{d})$. Otherwise, all subvectors of subgradients of $\psi$ over $\eta$ is zero except
%\begin{eqnarray}
%\partial_{\eta_{y_{d}}}\psi(y_{d},x_{d},\eta)=cx_{i}~~~\partial_{\eta_{\hat{y}}}\psi(y_{d},x_{d},\eta)=-cx_{i}.
%\end{eqnarray}
Using this subgradient in the SSGLD (Eqn.~\ref{eqn:SSGLD}) or SSGNHT (Eqn.~\ref{eqn:SSGNHT}), we can derive the stochastic subgradient inner sampler for classifiers $\eta$.

The whole stochastic HMC(LMC)-within-Gibbs algorithm structure is outlined in Algorithm.~\ref{alg:hmc within gibbs}.

\begin{algorithm}[H]
   \caption{Stochastic HMC within Gibbs for infinite SVM}
   \label{alg:hmc within gibbs}
\begin{algorithmic}
   \STATE {\bfseries Input:} data $(x_{d},y_{d}),d=1,\cdots,N$, batchsize $\tilde{N}$.
   \STATE Initialization
   \REPEAT
   \STATE sample $z$ given $\eta$
   \STATE sample $\eta$ given $z$ using stochastic subgradient HMC
   \UNTIL{Converge}
\end{algorithmic}
\end{algorithm}
\section{Experiments}\label{sec:experiment}
We now implement our stochastic subgradient MCMC on various Bayesian max-margin models, including the basic Bayesian linear SVM and two sophisticated Bayesian max-margin models with latent variables (GiSVM and Gibbs MedLDA).
Our results demonstrate that stochastic subgradient MCMC can achieve great improvement on time efficiency and meanwhile still generating accurate posterior samples.
%being accurate in posterior samples.
%cases for each model to get a comparison. are considered to show the comparison.

All experiments are done on a desktop computer with single-core rate up to 3.0GHz. The stepsize parameter at iteration t decays via $h_{t}=h_{0}*(1+t/b)^{-\gamma}$. Normally, we set $b=1$ for SVM classifier $\eta$ and $b=100$ for topic-word parameter $\Phi$. We choose $h_0$ and $\gamma$ via a grid search. Furthermore, the AdaGrad stepsizes are considered for stochastic subgradient Langevin dynamics method.

\subsection{Bayesian Linear SVMs}
We first consider the basic Bayesian linear SVM model and compare our stochastic subgradient sampling methods with the Gibbs sampler with data augmentation~\citep{zhu14medlda} and the random walk Metropolis with stochastic MH test~\citep{korattikara2014austerity} (stochastic random walk Metropolis, SRWM).

\subsubsection{Results on Synthetic Data}
We first test our methods on a 2D synthetic dataset to show that our methods give correct samples from the posterior distribution. Note that we view the results of this experiment as a simple proof of idea and hence choose the more direct visual comparison.
We follow the Bayesian linear SVM model defined in Section~\ref{sec:subgradienthmc} and generate 1000 observations as the synthetic dataset. Specifically, we generate features $x$ from a uniform distribution $x_i \stackrel{i.i.d} \sim U(0,1)$ and the coefficient vector from a normal distribution $\eta \sim \gauss(0, 1/3\cdot I)$. Given the features and coefficients, the labels are generated from the Bernoulli distribution with parameter $\delta$, where,
$$\delta=\frac{ \psi(y_i=1 | x_{i}, \eta) }{\psi(y_i=1 | x_{i}, \eta) + \psi(y_i=-1 | x_{i}, \eta)}.$$

\begin{figure}[ht]\vspace{-.2cm} \centering
{
\includegraphics[width=.7\columnwidth]{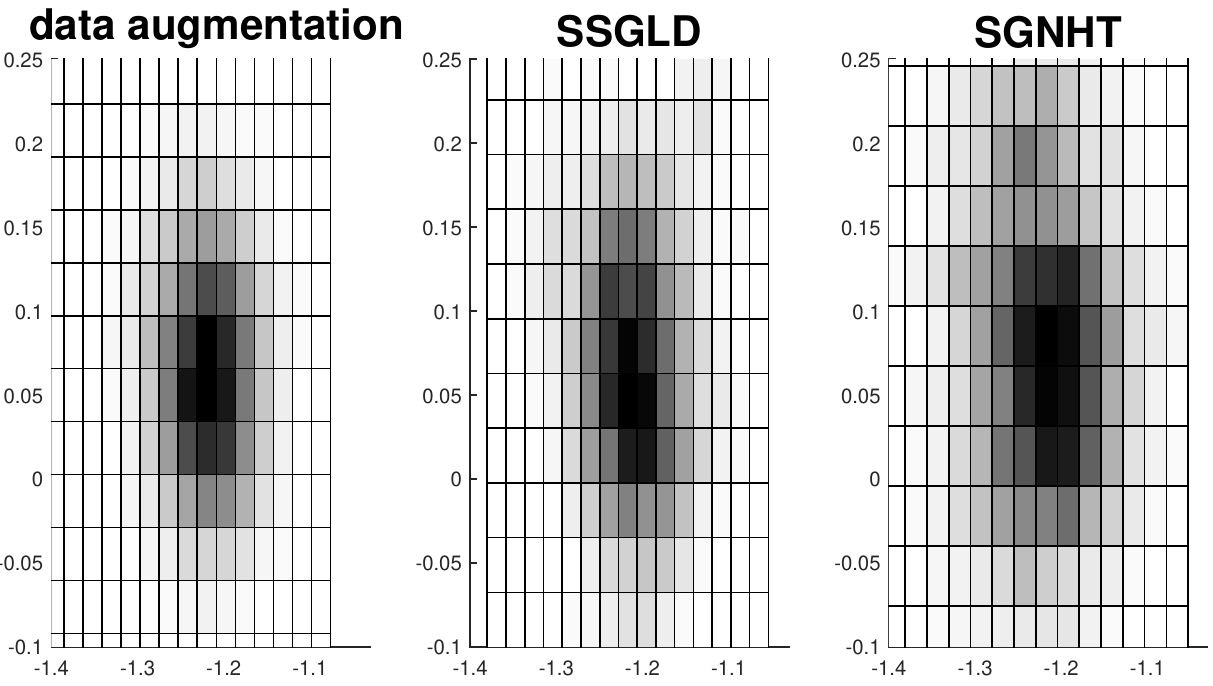}
}
\vspace{-6pt}
\caption{Visual comparison of posterior samples}
\vspace{5pt}
\label{fig:posterior contrast}
\end{figure}

We compare the samples obtained from SSGLD and SSGNHT with those from the data augmentation method which is an accurate sampler for Bayesian SVMs.
We take 5,000 samples for each method after a sufficiently long burn-in stage and give the comparison in Fig.~\ref{fig:posterior contrast}, where the densities of the obtained samples are shown via the grayscales of the grids.
The results suggest that our stochastic subgradient MCMC methods are accurate, although the stochastic subsampling and the neglect of MH test bring some noise.
This result is compatible with the previous weak convergence analysis of the ordinary HMC methods~\citep{sato2014sgldconvergence,chen2015splitting}.

\subsubsection{Results on Real Data}
We then test two stochastic subgradient MCMC methods, SSGLD and SSGNHT on the Realsim dataset~\footnote{\url{http://csie.ntu.edu.tw/~cjlin/libsvmtools/datasets/binary.html}} and the larger UCI Higgs dataset~\citep{asuncion2007uci}. The Higgs dataset contains $1.1\times 10^7$ samples in a 28-dimensional feature space. We randomly choose $10^7$ samples as the training set and the rest as the testing set.

\begin{figure}[t]\vspace{-.1cm} \centering
{
\includegraphics[width=.8\columnwidth]{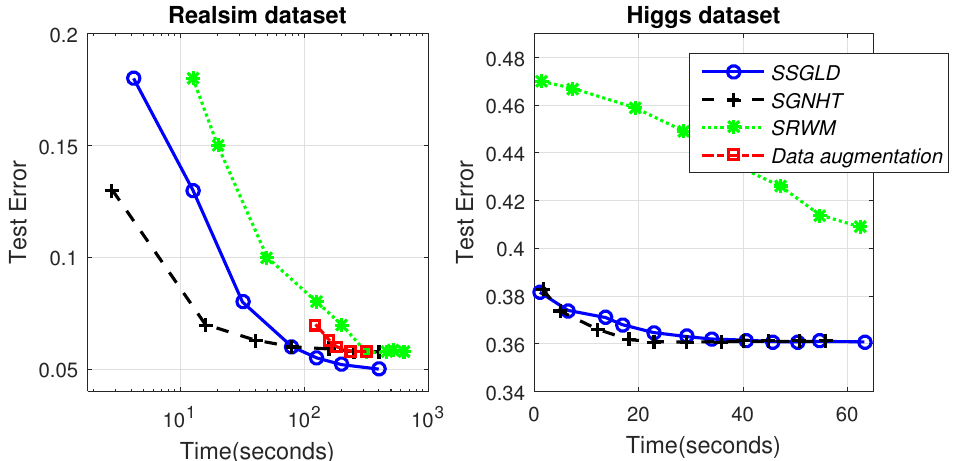}
}
\vspace{-6pt}
\caption{Experimental results of Bayesian linear SVMs}
\vspace{-10pt}
\label{fig:svm_Higgs_Realsim}
\end{figure}

For the Realsim dataset, we set the stochastic batchsize $|\tilde{\data}|=10$ for all stochastic inference methods. For Higgs dataset, we set $|\tilde{\data}|$ to be $1,000$ for both SSGLD and SRWM and $|\tilde{\data}|=100$ for SSGHNT.
We use tuned polynomial decaying stepsizes for stochastic subgradient MCMC methods and specifically for SSGLD, we prefer adaptive stepsize AdaGrad, which has been successfully applied in the stochastic (sub)gradient descent~\citep{duchi2011adaptive}. For SRWM, the variance parameter is set as $0.01$. These turn to be a good setting analyzed in the following sensitivity analysis in Section~\ref{sec:sensitivity}.

The convergence curves of various methods with respect to the running time on both datasets are shown in Fig.~\ref{fig:svm_Higgs_Realsim}. We can see that our stochastic subgradient MCMC methods are several magnitudes faster than the baseline methods. Compared with the Gibbs sampling with data augmentation method, stochastic subgradient MCMC methods get much cheaper updates and hence are more scalable. Specially for the larger Higgs dataset, a single update of Gibbs sampling is not finished when the stochastic subgradient MCMC get converged.  Furthermore, although both SRWM and stochastic subgradient MCMC use stochastic minibatches, stochastic subgradient MCMC methods mix much faster than SRWM because the posterior subgradient information provides the right direction to the true posterior.

\subsubsection{Sensitivity Analysis}\label{sec:sensitivity}
\begin{figure}[ht] \centering
{
\includegraphics[width=.9\columnwidth]{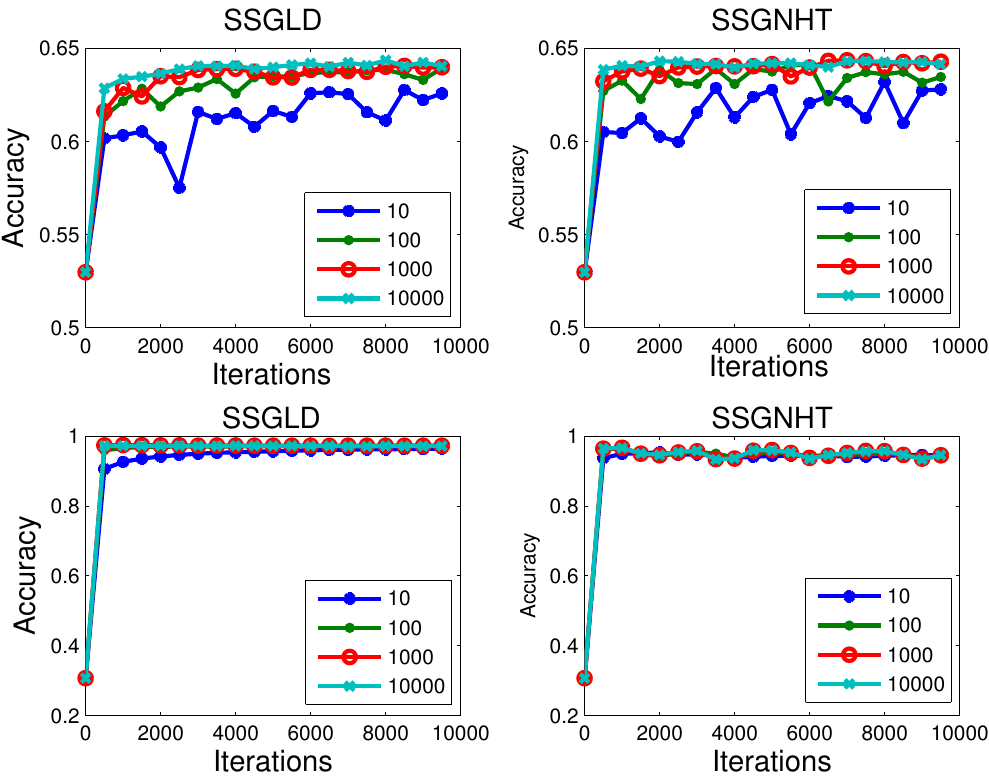}
}
\vspace{-6pt}
\caption{Sensitivity analysis of the batchsize parameter for both SSGLD and SSGNHT on the Higgs dataset ({\bf first row}); and the Realsim dataset ({\bf second row}).}\label{fig:svmbatchsize}
\vspace{-5pt}
\end{figure}

Tuning the batchsize $|\tilde{\data}|$ reflects an accuracy-efficiency trade-off,
analogous to the bias-variance tradeoff in stochastic Monte Carlo sampling~\citep{korattikara2014austerity}.
In general, using a smaller batchsize often leads to a larger injected noise, but the computation cost at each iteration
is reduced, which is linear to the batchsize (i.e., $O(|\tilde{\data}|$).
When doing cross validation to select parameters, both accuracy and time efficiency are key factors that should be taken into consideration.

Fig.~\ref{fig:svmbatchsize} presents the sensitivity analysis of the batchsize for the two stochastic subgradient MCMC methods on both Higgs and Realsim datasets.  The performance of our stochastic subgradient MCMC appears to be fairly promising except for extremely tiny batchsizes.

\begin{wrapfigure}{r}{0.4\textwidth}
\vspace{-10pt}
  \begin{center}
    \includegraphics[width=0.4\textwidth]{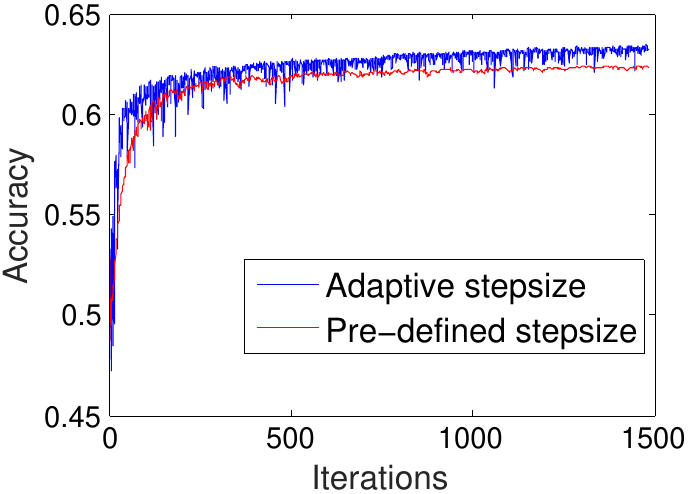}%{./Pictures/mainscreen1.png}
      \vspace{-10pt}
    \caption{Performance of SSGLD with AdaGrad}
    \label{fig:adagrad}
  \end{center}
  \vspace{-35pt}
  \vspace{0pt}
\end{wrapfigure}

In our experiments, adaptive stepsizes (AdaGrad) bring a better mixing rate than the polynomial decaying stepsizes. This may result from the flexible stepsize decaying at different dimensions.
We also give an empirical analysis in Fig.~\ref{fig:adagrad}. As can be seen, for the Higgs dataset, adaptive stepsizes bring better classification results than the pre-defined polynomial-decaying stepsizes.
%We also give an empirical analysis on adaptive stepsizes on the Higgs dataset in Fig.~\ref{fig:adagrad}.

\subsection{Gibbs max-margin Topic Models}
 Now, we implement the fast sampling for Gibbs MedLDA. We show the efficiency and accuracy of our stochastic subgradient Riemannian Langevin Dynamics (SSGRLD) using the 20news dataset and the larger Wikipedia dataset. Following the dataset setting in~\citep{zhu14medlda}, the stop words are removed according to a standard list. We compare our SSGRLD with the data augmentation (Gibbs MedLDA)~\citep{zhu14medlda} and its newly developed extension in the online Bayesian passive-aggressive learning framework (paMedLDA-gibbs)~\citep{shi2014onlinebayespa}. For the smaller 20news dataset, the involved three methods all use the binary version and then adopt the ``one-vs-all'' strategy for multi-class classification. For the larger Wikipedia dataset, the SSGRLD method uses the multi-class setting and other two use the multi-task formulation as described in~\citep{zhu14medlda,shi2014onlinebayespa}.

\subsubsection{Classification Performance}
\begin{figure}[ht]\centering
{
\includegraphics[width=.7\columnwidth]{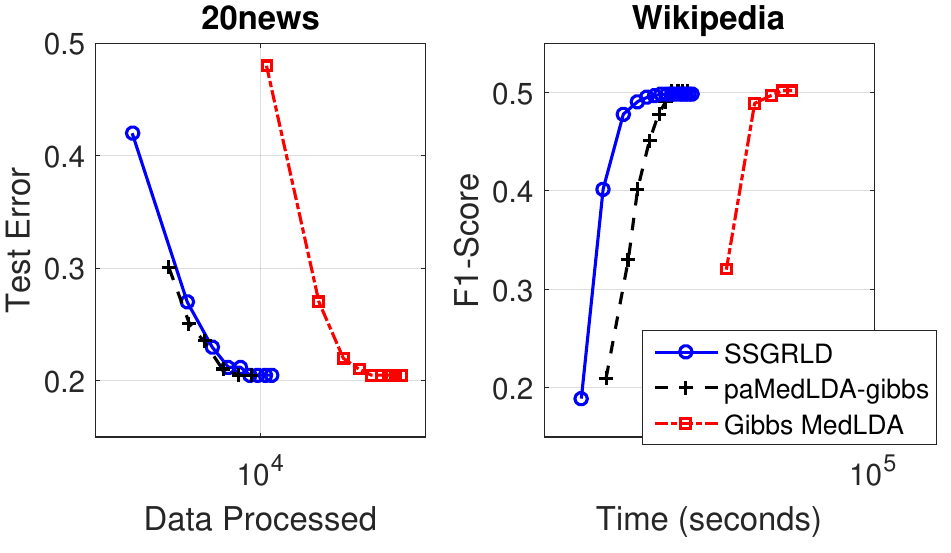}
}
\vspace{-10pt}
\caption{Empirical results of different methods for MedLDA}
\label{fig:medlda}
\vspace{-10pt}
\end{figure}

We first test on the 20news dataset which consists of 11,269 training documents and 20 categories. We set the hyper-parameters as $\alpha=1,\beta=1,c=1,\ell=164$ as suggested in~\citep{zhu14medlda}. Fig.~\ref{fig:medlda}(left) shows the number of documents processed in order to reach a specific accuracy score, where topic number is set as 50.
%needed for the document classification,
 As we can see, the two stochastic samplers use much fewer documents and efficiently explore the data redundancy by using a minibatch at each iteration.

 Then we test on the larger Wikipedia dataset which consists of 1.1 million training documents and 20 categories.  We use the same hyper-parameter setting with the 20news dataset, except for a few settings: $c=10, \ell=196$ for SSGRLD and $\ell=1$ for both Gibbs MedLDA and paMedLDA-gibbs. We set the topic number as 40. Fig.~\ref{fig:medlda} shows the F1-scores as a function of time. It can be seen that SSGRLD produces comparable classification results. As for the efficiency, both SSGRLD and paMedLDA-gibbs are one order of magnitude more efficient than the previous Gibbs MedLDA. This is due to the minibatch training. Meanwhile, although in the same magnitude, SSGRLD is still faster than paMedLDA-gibbs. We argue that this is because SSGRLD does not use augmented variables and directly draws samples from the SVM classifier. Moreover, the matrix inversion involved in the data augmentation technique is costly in the whole procedure.

\subsubsection{Topic Representations}
\begin{figure}[ht]\vspace{-10pt} \centering
{
\includegraphics[width=.7\columnwidth]{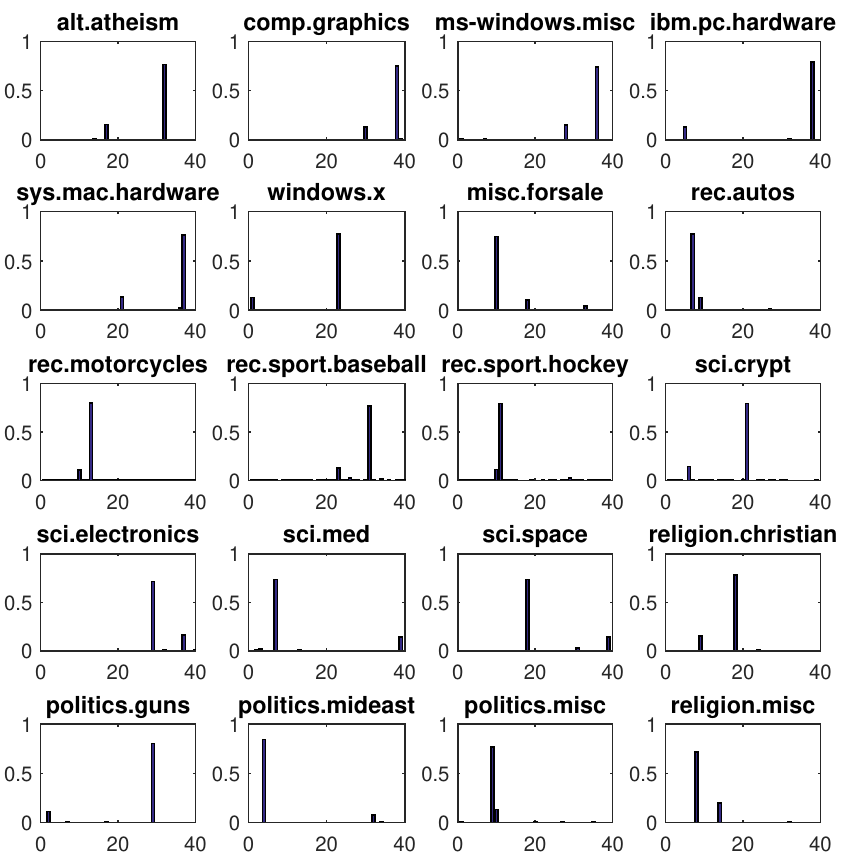}
}
\vspace{-0pt}
\caption{Visualization of learnt topics by SSGRLD}
\label{fig:topicvisualization}
\vspace{-1pt}
\end{figure}

Finally, we visualize the discovered topic representations of SSGRLD on the 20news dataset. For the all 20 categories, we show the average topic representations of the documents form each category. As we can see in Fig.~\ref{fig:topicvisualization}, the average topic distribution for the corresponding classifier is very sparse (only one or two non-zero entries). We also give the most representative top words of the salient topic(s) of each category in Table.~\ref{table:topwords}. We can see that the top words of the salient topic(s) are highly related to the category information.
For example, the salient topic learned by classifier \emph{sci.space} has the top words as NASA, launch, moon, satellite, etc.
%The salient topic learned by classifier \emph{politics.mideast} has the top words as Israel, Turkish, Armenian, killed, Greek, Arab, etc.
%This topic sparsity and structure are similar to the results of the previous MedLDA studies~\citep{zhu14medlda,shi2014onlinebayespa}
These patterns are similar as those in ~\citep{zhu14medlda,shi2014onlinebayespa}.
\begin{table}[ht]
\vspace{-15pt}
\caption{Representative top words of the salient topic(s)}
\label{table:topwords}
\begin{center}
\setlength\tabcolsep{3pt}
\begin{tabular}{cc|cc}
\hline\hline
Category & Top words & Category & Top words \\
\hline
atheism  & god, don, atheism & graphics  & image, jpeg, file\\
windows  & windows, file, card & pc  & scsi, drive, disk, mb, dos\\
mac  & mac, apple, drive  & windows  & window, server, file\\
forsale  & anonymity, sphinx & rec.autos  & car, engine, speed\\
motocycle  & bike, ride, bmw & baseball  & team, game, runs\\
hockey  & team, nhl, season & crypt  & key, chip, security, law\\
electronics  & power, circuit, wire &medical  & food, medical, doctor\\
space  & nasa, launch, earth & christian  & god, jesus, church, bible\\
 guns & gun, weapon, firearm & mideast  & israel, turkish, jews, arab\\
politics  & mr, president, states & religion  & jesus, bible, christian\\
\hline\hline
\end{tabular}
\end{center}
\vspace{-15pt}
\end{table}

\subsection{Infinite SVMs}
The proposed subgradient-based sampling methods can also be used for fast inference of infinite SVM~\citep{zhu2011infinite}, a Dirichlet process mixture of large-margin kernel machines.

We choose two datasets, Protein and IJCNN1, to test our methods. The Protein dataset~\citep{zhang2014max} was created for Protein fold classifications and consists of 698 samples and 27 classes with 21 features. The IJCNN1 dataset\footnote{\url{http://csie.ntu.edu.tw/~cjlin/libsvmtools/datasets/binary.html}} is originated from an engine system binary classification problem and consists of 49,990 training samples with 22 features.

We implement two inference methods for iSVM including SSGNHT within Gibbs (Algorithm.~\ref{alg:hmc within gibbs}) and Gibbs sampling with data augmentation~\citep{zhang2014max}. Other models are also implemented for comparison, such as multinomial logit model (MNL), linear SVM, RBF-SVM and DP mixture of generalized linear models (dpMNL)~\citep{shahbaba2009dpmnl}. We use cross-validations to choose hyper-parameters and get the results in Table.~\ref{table:isvm}.

We can see that nonlinear models using a mixture-of-experts, such as GiSVM and dpMNL, are superior in classification. In the stochastic subgradient MCMC, $\eta$ sampling step can be dramatically accelerated, with comparable or even better prediction performance. This superiority results from both stochastic subsampling and avoiding the matrix inversion in the data augmentation technique.

\begin{table*}[t]
\caption{Efficiency (in minutes) and accuracy of various models on the Protein and IJCNN1 datasets}
\vspace{-15pt}
\label{table:isvm}
\begin{center}
\setlength\tabcolsep{3pt}
\begin{tabular}{c|ccc|ccc}
\hline\hline
Datasets & \multicolumn{3}{c}{Protein} & \multicolumn{3}{|c}{IJCNN1} \\
 & Accu(\%)  & Time for $\eta$ & Total Time &Accu(\%) & Time for $\eta$ & Total Time\\
\hline
MNL & {50.0}  &  -   & 0.10 &${91.3}$  &  -   & 3.21\\
Linear SVM & {50.8}  &  -   & 0.03 & ${91.0}$  &  -   & 0.56 \\
RBF-SVM & {53.1}  &  -   & 0.11&${93.9}$  &  -   &  2.79\\
dpMNL & \textbf{{56.3}}  &  -   & 7.64 & 94.0 &  -   & 7.62\\
\hline
Gibbs-iSVM & 55.8$\pm$0.0  &  8.31$\pm$0.27    & 15.15$\pm$0.29 & \textbf{94.2$\pm$0.7}  &  $9.13\pm0.95$   & 22.71$\pm$1.16\\
SSGNHT-iSVM &   56.1$\pm$0.0  & \textbf{0.17$\pm$0.02}   & \textbf{7.32$\pm$0.26} &  \textbf{94.2$\pm$0.8}  & \textbf{1.17$\pm$0.08}  & \textbf{13.84$\pm$1.90}\\
\hline\hline
\end{tabular}
\end{center}
\vspace{-15pt}
\end{table*}

\section{Conclusions}
% summary of paper
We systematically investigate the fast sampling methods for Bayesian max-margin models.
We first study a general subgradient HMC sampling method and several stochastic variants including SSGLD and SSGNHT. Theoretical analysis shows the approximated detailed balance of the proposed stochastic subgradient MCMC methods.
Then we apply the stochastic subgradient samplers to Bayesian linear SVMs and two sophisticated Bayesian max-margin models with latent variables (GiSVM and Gibbs MedLDA).
Extensive empirical studies demonstrate the effectiveness of the stochastic subgradient MCMC methods on improving time efficiency while maintaining a high accuracy of the samples.

% concludes the strengths and weaknesses
The strengths of our methods are 1) fast inference for BMM models compared with the previous Gibbs sampling method with data augmentation; 2) accurate sampling which is as good as the Gibbs sampling with data augmentation and 3) applications to non-conjugate posterior sampling which cannot be simply accomplished.
However, when the data sizes of the applications are too large to be processed in a single machine, it is still difficult to use only stochastic subgradient MCMC to solve the problem.
%Our methods could be extended to even larger scales by using the parallel computation techniques~\citep{ahn2014distributed}.

% future research direction
We consider the future work in three categories: algorithm-level, model-level and application-level. For the proposed algorithm itself, the future work includes further scaling up using parallel computation~\citep{ahn2014distributed}. For the model setting, the future work includes applying our method to other models with continuous but non-smooth posteriors, such as sparse models with Laplacian priors. At the application level, we consider using our method to scale up several Bayesian max-margin models that are used in intelligent systems, such as nonparametric max-margin matrix factorization for collaborative filtering~\citep{xu2012nonparametric}.

% Applications and impacts
%Our fast sampling method potentially scale up the future applications of Bayesian max-margin models, including text categorization~\citep{zhu2012medlda}, collaborative filtering~\citep{xu2012nonparametric}, social network prediction~\citep{zhu2012max} and crowdsourcing~\citep{tian2015max}, which are the major problems in real intelligent systems.
%contribution in experting and intelligent systems
The big data is identified as an important building block of intelligent systems~\citep{manyika2011big,lehmann2015dbpedia} and the fast inference is becoming a central element therein~\citep{park2015reversed}. For related Bayesian models~\citep{pearl2014probabilistic}, big learning with Bayesian models is one of the recent research focuses~\citep{zhu2014big}.
Particularly, the Bayesian max-margin models are well studied for various machine learning applications, but they still lack fast inference methods. Our method accomplishes fast sampling for the BMM models, which will be used in future large scale intelligent systems.

\bibliography{ref}

\end{document}